\documentclass[lettersize,journal]{IEEEtran}
\usepackage{amsmath,amsfonts}
\usepackage{amssymb}
\usepackage[linesnumbered,ruled,vlined]{algorithm2e}
\usepackage{array}
\usepackage[caption=false,font=normalsize,labelfont=sf,textfont=sf]{subfig}
\usepackage{textcomp}
\usepackage{stfloats}
\usepackage{url}
\usepackage{verbatim}
\usepackage{graphicx}
\usepackage{colortbl}
\usepackage{placeins}
\usepackage{booktabs}
\usepackage{multirow}
\usepackage{cite}
\usepackage{adjustbox}
\usepackage{hyperref}
\usepackage{tabularx}
\usepackage{makecell}
\hypersetup{
    colorlinks=true,      
    linkcolor=blue,       
    citecolor=blue,       
    urlcolor=blue         
}
\usepackage{xcolor}
\usepackage{tcolorbox}
\usepackage{pifont}

\begin{document}

\title{GTSM: Gaussian Splatting with Thermal State-Space Dynamics for Spatio-Temporal Thermal Field Reconstruction and Prediction}

\author{Yunpeng Zhao,~\IEEEmembership{Graduate Student Member,~IEEE}, Haoxiang Zhang, and Chao Liu
\thanks{Yunpeng Zhao and Haoxiang Zhang are with the Key Laboratory for Information Science of Electromagnetic Waves (MoE), Fudan University, Shanghai 200433, China  (e-mail:ypzhao23@m.fudan.edu.cn){(corresponding author: Yunpeng Zhao)}}%
\thanks{Chao Liu is with the Department of Electronic Engineering, Tsinghua University, Beijing, China (e-mail: chao-liu23@mails.tsinghua.edu.cn).} }%

\markboth{Journal of \LaTeX\ Class Files,~Vol.~14, No.~8, August~2021}%
{Shell \MakeLowercase{\textit{et al.}}: A Sample Article Using IEEEtran.cls for IEEE Journals}

\IEEEpubid{0000--0000/00\$00.00~\copyright~2021 IEEE}

\maketitle

\begin{abstract}
Temperature-field reconstruction and prediction are important for condition monitoring, process analysis, and state assessment in industrial and dynamic environments. This work aims to recover a four-dimensional spatio-temporal thermal state field directly from two-dimensional observations, without requiring a predefined 3D model, prescribed thermodynamic equation, or predefined heating/cooling trajectory. To this end, we propose the Gaussian Thermal State-Space Model (GTSM), which represents thermal evolution through a shared scene-level latent process and Gaussian-specific local thermal states. The resulting continuous dynamics are propagated using differentiable RK2 integration, allowing thermal evolution to be inferred from multi-view and multi-temporal observations and continuously extrapolated beyond the observed interval. GTSM is designed as a backbone-agnostic thermal dynamics module for Gaussian representations and can be integrated with both static 3D Gaussian Splatting and dynamic 4D Gaussian Splatting, supporting thermal reconstruction and prediction under both static and time-varying target geometry. The reconstructed representation further enables novel-view thermal rendering and future-time thermal-state prediction. Experiments on real infrared sequences and synthetic dynamic thermal data demonstrate improved reconstruction accuracy and substantially stronger temporal extrapolation than representative baselines, while also validating the applicability of GTSM to deforming targets.
\end{abstract}


\section{Introduction}

\IEEEPARstart{T}{emperature} is an important physical indicator for characterizing the operating state, energy dissipation, and potential anomalies of industrial equipment and dynamic targets~\cite{OZ10682482}. Accurate temperature field reconstruction has therefore been widely applied to equipment condition monitoring~\cite{WANG2022105359}, combustion analysis~\cite{YANG2026128592}, and infrared target perception~\cite{YAMASHITA2026128398}. In many practical scenarios, however, a temperature field at a single instant provides only limited state information. Its temporal evolution often reflects changes in heat input, operating conditions, and heat-transfer processes. Therefore, reconstructing not only the current temperature distribution but also its temporal evolution is important for understanding target states and predicting their future thermal behavior.

\begin{figure}[t]
    \centering
    \includegraphics[width=\columnwidth]{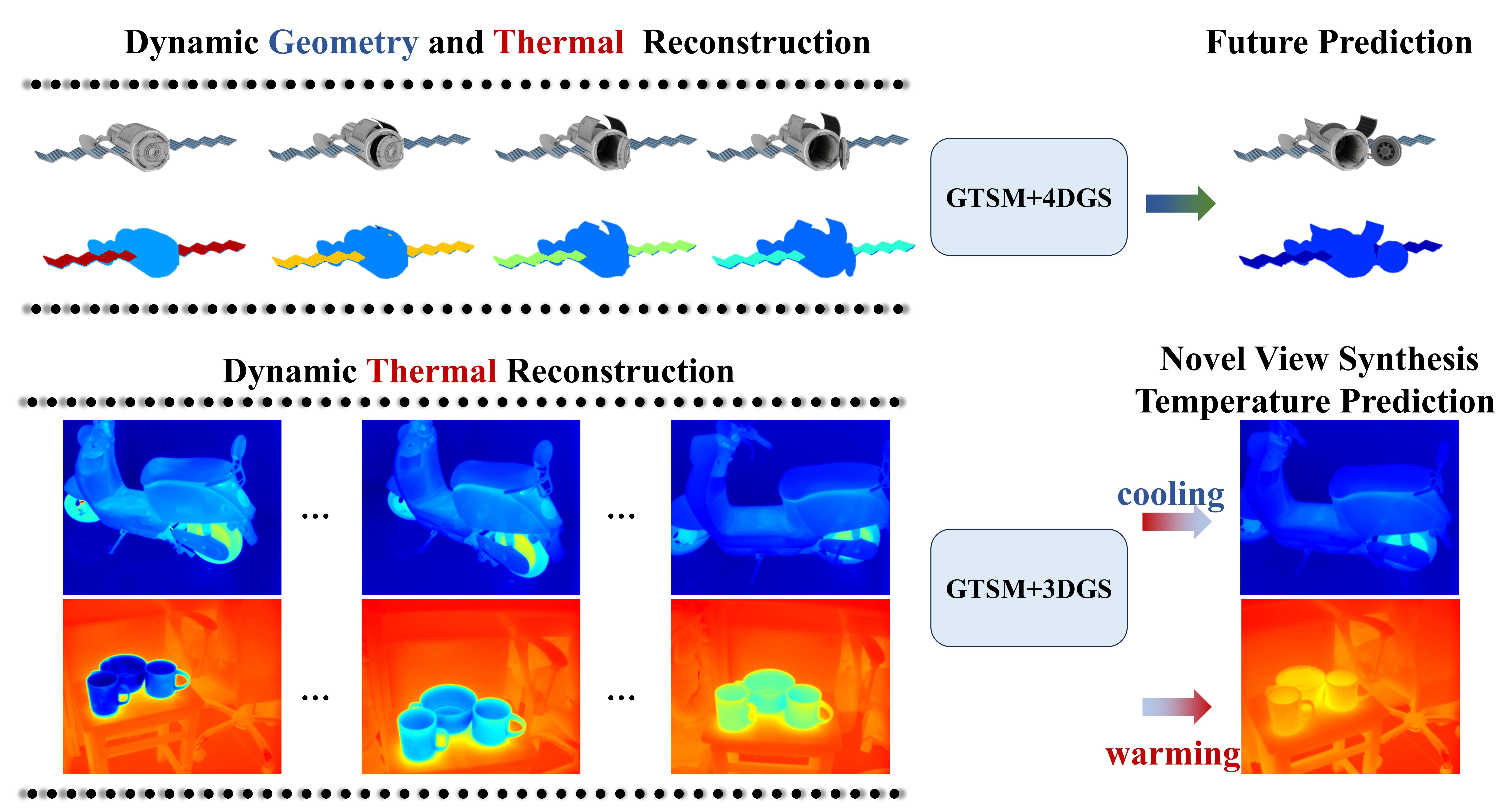}
    \caption{GTSM reconstructs spatio-temporal thermal fields from two-dimensional observations and enables novel-view synthesis and future thermal-state prediction for both static and dynamic targets.
    }
    \label{fig:task_overview}
\end{figure}

Traditional thermal reconstruction and prediction methods typically formulate temperature evolution using heat-transfer equations and solve the resulting physical models through numerical techniques such as finite-element or CFD-based simulation~\cite{YANG2020111448}. These methods can provide physically consistent solutions when the target geometry, material properties, heat sources, and boundary conditions are known. Data-driven methods, including long short-term memory networks (LSTMs)~\cite{FANG2025111170} and Transformer-based models~\cite{wu2024transolver}, provide an alternative way to learn temperature evolution directly from observations. physics-informed/physics-guided learnings~\cite{LU2026128309,gong_tfrd, Hadzhamolla_2026} further introduce physical constraints or continuous dynamics into thermal prediction. Nevertheless, most of these approaches operate on predefined
spatial domains, meshes, or image coordinates, and generally assume that the spatial correspondence of the temperature field is already known.

When an accurate geometric model and thermal-process prior are unavailable, thermal reconstruction becomes a substantially more difficult computational inverse-imaging problem. In many practical scenarios, the available measurements consist only of two-dimensional infrared or temperature observations acquired at different times and viewpoints, while the underlying three-dimensional geometry and spatio-temporal thermal states are unknown. The challenge is therefore to recover both the spatial representation and its time-varying thermal state directly from these image measurements, without relying on a predefined mesh, known thermophysical coefficients, or a prescribed temperature-evolution law.

\IEEEpubidadjcol

Differentiable neural rendering provides a natural computational imaging framework for addressing the geometric part of this inverse problem. 3D Gaussian Splatting (3DGS) has attracted considerable attention because of its explicit primitive representation, efficient optimization, and high-quality differentiable rendering~\cite{Kerbl}. By introducing time-dependent deformation or dynamic Gaussian attributes, 4DGS methods further extend this capability to scenes and targets whose geometry changes over time
\cite{Wu_2024_CVPR,yang2023deformable3dgs}. These developments provide an explicit and differentiable image-formation model for recovering either static or dynamic three-dimensional geometry from multi-view image observations.

However, dynamic thermal-field reconstruction remains much less explored. Compared with conventional geometric reconstruction, this problem involves two coupled inverse processes. The first is the spatial inversion from two-dimensional observations to a three-dimensional target representation. The second is the recovery of a thermal state on this representation, requiring spatial consistency between temperature and surface primitives as well as temporal consistency across observations. The ultimate goal is therefore to recover a four-dimensional spatio-temporal thermal state field from two-dimensional measurements.

This problem is particularly challenging because the underlying thermal process is generally unknown, while different surface elements can exhibit distinct temporal responses due to variations in material properties, spatial location, thermal exposure, and local boundary conditions. A straightforward strategy is to directly regress the temperature of each Gaussian as a function of time using a time-conditioned MLP or a generic temporal network. However, such formulations mainly describe the observed temperature trajectory and do not explicitly distinguish between the thermal process shared by the scene and the heterogeneous local responses of individual surface primitives. This lack of structured thermal dynamics can make it difficult to capture spatially heterogeneous temporal responses and to robustly propagate the learned state beyond the observed temporal interval.

To address these challenges, we propose the \textbf{Gaussian Thermal State-Space Model (GTSM)}, a structured Gaussian-level thermal dynamics representation for spatio-temporal thermal-field reconstruction and prediction. Instead of independently regressing the temporal temperature trajectory of each Gaussian, GTSM decomposes thermal evolution into a scene-level latent thermal process and Gaussian-specific local thermal states. The scene-level latent process captures the shared temporal thermal driving pattern of the target, while the Gaussian-specific states model spatially heterogeneous responses to this common process.

The coupled thermal states evolve through continuous differential dynamics and are jointly inferred from multi-view and multi-temporal observations through differentiable Gaussian rendering. Importantly, GTSM does not require a prescribed thermodynamic equation, known thermophysical coefficients, or a predefined heating or cooling trajectory. Instead, it learns a structured continuous thermal dynamics model directly from observations. This formulation enables the reconstructed Gaussian representation not only to explain the observed spatio-temporal thermal field, but also to propagate the learned thermal state beyond the observation interval for future-time thermal prediction.

The main contributions of this work are summarized as follows:

\begin{itemize}

    \item We propose GTSM for reconstructing four-dimensional spatio-temporal thermal fields from multi-view and multi-temporal two-dimensional observations. GTSM augments Gaussian primitives with continuous thermal states represented through a shared scene-level latent process and Gaussian-specific local dynamics.

    \item We introduce continuous Gaussian-level state-space dynamics that infer temporal thermal evolution directly from observations without prescribing a specific heating or cooling trajectory. The learned dynamics support both reconstruction within the observed interval and continuous rollout for future-time prediction.

    \item GTSM is designed as a backbone-agnostic thermal dynamics module for Gaussian representations and can be integrated with both static 3DGS and dynamic 4DGS frameworks. Experiments on real infrared sequences and synthetic dynamic thermal data validate its effectiveness for within-interval reconstruction, novel-view rendering, and future-time thermal prediction.

\end{itemize}

The remainder of this paper is organized as follows. Section~\ref{sec2} reviews related work. Section~\ref{sec3} presents the proposed method in detail. Section~\ref{sec5} describes the experimental setup, reports results, and includes ablation studies. Finally, Section~\ref{sec5} concludes the paper.

\section{Related work}
\label{sec2}

\subsection{Static and Dynamic Scene Reconstruction with Gaussian Splatting}

3D Gaussian Splatting (3DGS)~\cite{Kerbl} represents a scene using explicit anisotropic Gaussian primitives and reconstructs the scene through differentiable rasterization. Its explicit representation, high rendering efficiency, and strong novel-view synthesis quality have motivated extensive follow-up research in scene reconstruction and related rendering tasks~\cite{GS2Physics,Recon}.

Subsequent studies extend 3DGS from static scenes to time-varying environments. Dynamic Gaussian methods, commonly referred to as 4DGS, introduce temporal deformation or time-dependent Gaussian representations to describe scene motion
and geometric changes. Representative approaches model temporal variation using deformation fields~\cite{yang2023deformable3dgs}, spatio-temporal
grids~\cite{Wu_2024_CVPR}, control-point-based transformations
~\cite{huang2024scgs,yang2023gs4d}, or explicit space-time Gaussian representations~\cite{duan:2024:4drotorgs}. These methods enable Gaussian primitives to evolve with time and therefore support high-quality reconstruction and novel-view rendering of dynamic scenes.

More recently, research has begun to move beyond reconstruction at observed timestamps toward learning the underlying temporal dynamics of scene motion. Methods such as TRACE~\cite{li2025trace} introduce continuous dynamics into Gaussian motion modeling, allowing the evolution pattern of geometric states to be inferred from temporal observations and further propagated to unseen future timestamps. This development extends dynamic Gaussian reconstruction from time-conditioned scene representation toward temporal state prediction.

Despite these advances, most existing dynamic Gaussian methods primarily model the evolution of geometry and visual appearance. Compared with geometric dynamics, explicitly attaching a thermal state to Gaussian primitives, learning
its temporal evolution from image observations, and propagating the learned thermal state to future timestamps remain relatively underexplored.

\subsection{Thermal Reconstruction and Dynamics with Gaussian Representations}

Early NeRF-based\cite{HASSAN2025103345, Lin10644336} and 3DGS- based methods\cite{Lu,Liu,Myeongseok,ZHAO2027133982, Zhao-01, WangYaoxing}reconstruct infrared scenes from RGB--thermal image pairs or thermal-only observations, enabling three-dimensional scene reconstruction and thermal novel-view synthesis. Considering that thermal infrared imaging is governed by physical processes different from visible-light image formation, subsequent studies further introduce infrared radiation physics into Gaussian representations. Thermal3D-GS~\cite{Chen} incorporates atmospheric transmission and thermal conduction effects to improve thermal infrared novel-view synthesis. TPI-GS~\cite{ZHAO2026106593} and PhysIR-Splat~\cite{gao2026physir} model temperature, emissivity, environmental radiation, atmospheric transmission, and reflected components within the Gaussian rendering process. 

Recent studies have begun to extend Gaussian representations toward time-varying thermal reconstruction. NTR-Gaussian~\cite{Yang} combines time-conditioned neural temperature prediction with thermodynamics-inspired differential constraints and numerical integration to model nighttime thermal evolution. ETGS~\cite{wangetgs} introduces explicit thermodynamic dynamics into individual Gaussians and describes their temperature evolution using a first-order heat-transfer ODE with an analytical closed-form solution, enabling efficient reconstruction at arbitrary timestamps. These studies represent important progress from static thermal appearance reconstruction toward dynamic thermal-state modeling. These studies represent important progress from static thermal appearance reconstruction toward dynamic thermal modeling. However, existing formulations either rely strongly on prescribed thermodynamic structures or directly parameterize temperature as a function of spatial and temporal observations, while explicit Gaussian-level thermal states with learnable continuous dynamics remain less explored.

Different from these approaches, our GTSM models the thermal state through a scene-level latent thermal process and Gaussian-specific local thermal responses. The temporal evolution is learned directly from multi-view and multi-temporal observations using continuous state-space dynamics, without requiring a predefined heating or cooling trajectory. Moreover, GTSM is independent of a specific geometric backbone and can be integrated with both 3DGS and 4DGS, enabling spatio-temporal thermal-state reconstruction and cross-time prediction for both static and dynamically deforming targets.

\section{Methodology}
\label{sec3}

\subsection{Problem Formulation and Overall Framework}
\label{sec:problem_formulation}

This work considers the reconstruction and prediction of a four-dimensional spatio-temporal thermal field from two-dimensional observations acquired at different timestamps and camera poses. Given a sequence of thermal or infrared observations $\{\mathbf{Y}_k,\mathbf{P}_k,t_k\}_{k=1}^{K}$, where $\mathbf{Y}_k$ denotes the observed temperature map or infrared image, $\mathbf{P}_k$ represents the corresponding camera parameters, and $t_k$ is the acquisition timestamp, our goal is to recover a three-dimensional Gaussian representation of the target together with its continuously evolving surface
thermal state.

We formulate the scene using a Gaussian geometry backbone together with an additional thermal state attached to each primitive. The Gaussian backbone reconstructs the three-dimensional spatial representation from the available
image observations, while GTSM augments each Gaussian with a time-varying temperature $T_i(t)$. This formulation applies to either fixed or moving cameras and to targets with either static or time-varying geometry. Specifically, the $i$-th Gaussian primitive at timestamp $t$ is defined as
\begin{equation}
    \mathcal{G}_i(t)
    =
    \left\{
    \mathbf{x}_i(t),
    \mathbf{s}_i(t),
    \mathbf{r}_i(t),
    \alpha_i,
    \mathbf{c}_i,
    T_i(t)
    \right\},
    \label{eq:v28_dynamic_thermal_gaussian}
\end{equation}
where $\mathbf{x}_i(t)$, $\mathbf{s}_i(t)$, and $\mathbf{r}_i(t)$ denote the Gaussian position, scale, and rotation, respectively. The opacity $\alpha_i$ and
appearance attribute $\mathbf{c}_i$ describe its rendering properties, while $T_i(t)$ denotes the Gaussian-level thermal state.

The geometric and thermal states play different roles in this representation. GTSM does not alter the geometric parameterization of the Gaussian primitive; instead, it augments each Gaussian with an additional scalar thermal state
$T_i(t)$. For a geometrically static target, $\mathbf{x}_i$, $\mathbf{s}_i$, and $\mathbf{r}_i$ remain time-invariant, whereas $T_i(t)$ can evolve continuously. For a dynamic target, the geometric attributes additionally become time-dependent through the adopted dynamic Gaussian backbone. In both
cases, the thermal state is modeled independently of the geometric evolution while remaining attached to the same Gaussian primitive.

Given a camera pose and timestamp, the thermal observation is obtained by rendering the Gaussian representation at the queried time:
\begin{equation}
    \hat{\mathbf{Y}}_k
    =
    \mathcal{R}_{T}
    \left(
    \left\{
    \mathcal{G}_i(t_k)
    \right\}_{i=1}^{N},
    \mathbf{P}_k
    \right),
    \label{eq:v28_thermal_rendering_formulation}
\end{equation}
where $\mathcal{R}_{T}(\cdot)$ denotes differentiable thermal or infrared rendering. The rendered observation $\hat{\mathbf{Y}}_k$ is compared with the corresponding measurement $\mathbf{Y}_k$, and the reconstruction error is back-propagated through the differentiable renderer to optimize the Gaussian representation and the associated thermal dynamics. In this way, the spatial scene representation and its time-varying thermal state are jointly constrained by multi-view and multi-temporal observations.

The resulting representation supports both within-interval novel-view/temporal thermal reconstruction and future-time prediction. Rather than assigning independent temperatures to individual timestamps, GTSM learns continuous Gaussian-level thermal dynamics from the observed sequence, allowing $T_i(t)$
to be queried at unseen viewpoints and timestamps, including times beyond the observed interval.

\begin{figure*}[htbp]
    \centering
    \includegraphics[width=\textwidth]{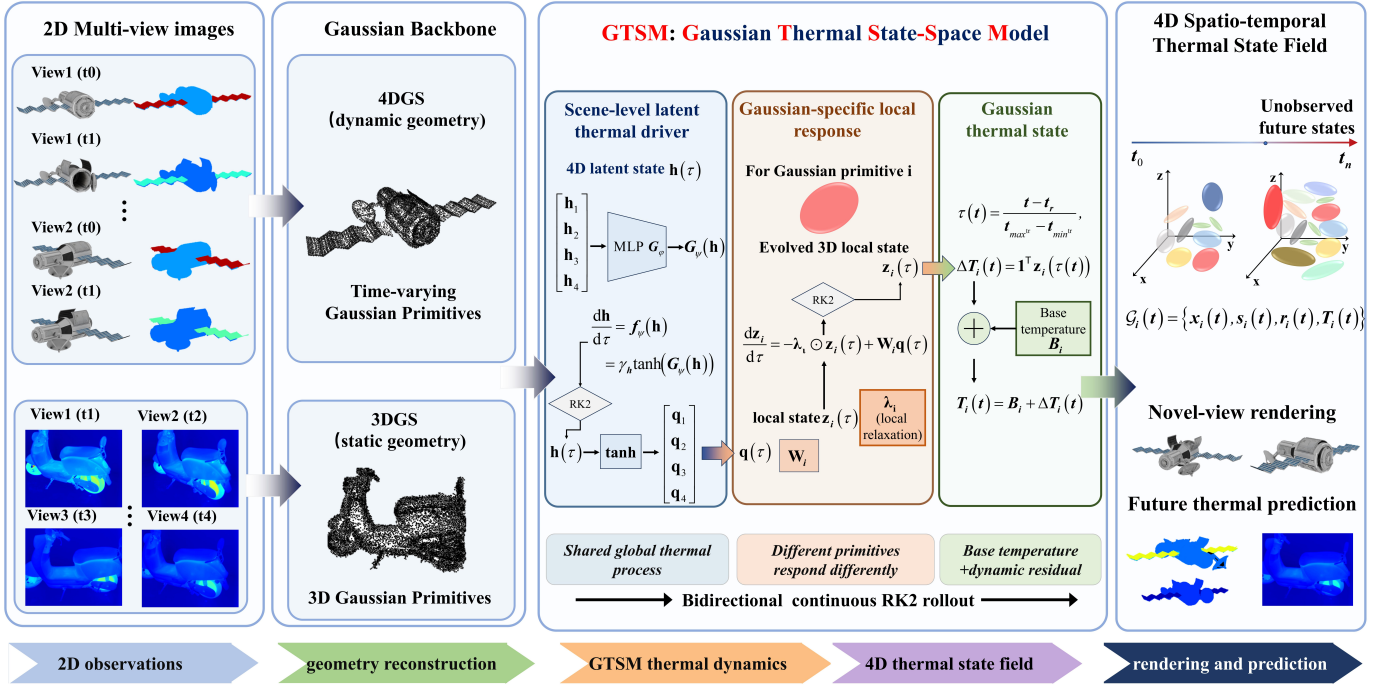}
    \caption{ Overview of the proposed GTSM framework. GTSM can be coupled with static 3DGS or dynamic 4DGS representations, and models Gaussian-level thermal evolution through a scene-level latent driver and Gaussian-specific local dynamics for spatio-temporal thermal field reconstruction, novel-view rendering, and future prediction.
    }
    \label{fig:stage3_detail}
\end{figure*}

Fig.~\ref{fig:stage3_detail} summarizes the overall framework. The Gaussian backbone provides the static or time-varying spatial representation, while GTSM models the attached thermal state $T_i(t)$ and its continuous temporal evolution.

\subsection{Gaussian-Level Local Thermal State-Space Model}
\label{sec:thermal_state_space}

Directly regressing \(T_i(t)\) with a generic time-conditioned network can fit observed temperature samples, but it does not explicitly distinguish the common
thermal process driving the scene from the heterogeneous responses of individual
surface primitives. In real thermal evolution, different surface elements may
respond differently to shared thermal excitation because of variations in
material properties, geometric exposure, spatial location, and local thermal
conditions. Moreover, the observed temperature generally contains a persistent
thermal level in addition to its time-varying component.

GTSM therefore introduces a structured thermal representation that separates
the persistent thermal component from its temporal variation and models the
latter through coupled scene-level and Gaussian-specific dynamics.

\textbf{Thermal base and dynamic residual representation.}
The temperature of the \(i\)-th Gaussian is represented as a persistent thermal
base plus a time-varying residual:
\begin{equation}
    T_i(t)
    =
    B_i
    +
    \Delta T_i(t),
    \label{eq:thermal_base_decomposition}
\end{equation}
where \(B_i\) denotes the learnable base temperature of the Gaussian and
\(\Delta T_i(t)\) describes its temporal variation. The base temperature is
initialized from the environmental temperature provided by the dataset metadata
and is subsequently optimized from the observations.

To describe temporal evolution consistently across sequences with different
durations, all thermal dynamics are formulated in a reference-centered
normalized temporal coordinate \(\tau\), defined as
\begin{equation}
    \tau(t)
    =
    \frac{t-t_r}
    {t_{\max}^{\mathrm{tr}}-t_{\min}^{\mathrm{tr}}},
    \label{eq:thermal_normalized_time}
\end{equation}
where \(t\) denotes the physical timestamp, while \(\tau\) is the normalized
temporal coordinate used for all subsequent state-space dynamics. The reference
time \(t_r\) and the corresponding rollout procedure are described later.

A single latent variable may be insufficient to represent thermal evolution
involving multiple response and relaxation modes. A \(D\)-dimensional local
thermal state \(\mathbf z_i(\tau)\in\mathbb R^D\) is therefore introduced for
each Gaussian, with the dynamic temperature residual defined as
\begin{equation}
    \Delta T_i(t)
    =
    \mathbf{1}^{\mathsf T}\mathbf{z}_i(\tau(t)).
    \label{eq:thermal_dynamic_residual}
\end{equation}
where \(\mathbf{1}^{\mathsf T}\) sums the \(D\) state channels to obtain the
scalar temperature residual. Rather than assigning a specific physical meaning
to each channel, \(\mathbf{z}_i(\tau)\) provides a compact multi-dimensional
representation of the time-varying thermal state. Its channels jointly determine
\(\Delta T_i(t)\), while their temporal evolution is governed by the scene-level
thermal driver and Gaussian-specific local dynamics described below.

\textbf{Scene-level thermal driving and Gaussian-level local response.}

To model the evolution of the local thermal state \(\mathbf{z}_i(\tau)\), GTSM separates the temporal dynamics into a shared scene-level driving process and Gaussian-specific local responses. The former captures the dominant thermal trend shared across the scene, while the latter accounts for heterogeneous responses of individual Gaussian primitives.

A low-dimensional scene-level latent state \(\mathbf{h}(\tau)\in\mathbb{R}^{M}\) is introduced to represent the shared thermal driving process. Its evolution is governed by an autonomous neural differential equation:
\begin{equation}
    \frac{\mathrm{d}\mathbf{h}}{\mathrm{d}\tau}
    =
    f_{\psi}\!\left(\mathbf{h}\right)
    =
    \gamma_h
    \tanh\!\left(
    G_{\psi}\!\left(\mathbf{h}\right)
    \right),
    \label{eq:global_thermal_driver}
\end{equation}
where \(G_{\psi}\) is a lightweight MLP and \(\gamma_h>0\) controls the
evolution rate of the latent state. The parameter \(\gamma_h\) is constrained
to be positive using a softplus parameterization. The corresponding
scene-level driving signal is defined as
\begin{equation}
    \mathbf{q}(\tau)
    =
    \tanh\!\left(\mathbf{h}(\tau)\right).
    \label{eq:thermal_driver_signal}
\end{equation}

Rather than prescribing a specific heating or cooling function, the shared thermal process is learned from the observed sequence through end-to-end optimization, providing a compact representation of the common temporal thermal dynamics of the scene.

The same thermal driver does not produce identical temperature changes at all surface locations. We therefore describe the local response of the $i$-th Gaussian as
\begin{equation}
    \frac{\mathrm{d}\mathbf{z}_i}{\mathrm{d}\tau}
    =
    -
    \boldsymbol{\lambda}_i
    \odot
    \mathbf{z}_i(\tau)
    +
    \mathbf{W}_i
    \mathbf{q}(\tau),
    \label{eq:local_thermal_state}
\end{equation}
where \(\boldsymbol{\lambda}_i\in\mathbb{R}_{+}^{D}\) contains the Gaussian-specific relaxation rates of the local thermal state, and \(\mathbf{W}_i\in\mathbb{R}^{D\times M}\) determines how strongly the Gaussian responds to the scene-level thermal driver. To ensure positive relaxation rates and stable dynamics, they are parameterized element-wise as \(\boldsymbol{\lambda}_i=\operatorname{softplus}(\hat{\boldsymbol{\lambda}}_i)+\epsilon\), where \(\hat{\boldsymbol{\lambda}}_i\) is unconstrained and \(\epsilon=10^{-5}\).

The two terms in Eq.~\eqref{eq:local_thermal_state} have different roles. The first term, $-\boldsymbol{\lambda}_i\odot\mathbf{z}_i$, describes the tendency of the local thermal residual to relax when no additional driving is applied. A larger \(\lambda_{i,d}\) produces faster relaxation of the corresponding state channel, whereas a smaller value results in slower decay and longer thermal memory. Therefore, \(\boldsymbol{\lambda}_i\) can be interpreted as a set of learnable relaxation rates in the normalized temporal domain, rather than directly measured physical coefficients. The second term, $\mathbf{W}_i\mathbf{q}(\tau)$, describes how the common thermal driving signal is converted into the local response of the Gaussian. The entries of \(\mathbf W_i\) determine how different components of the shared driver contribute to each local state channel. It therefore acts as an effective thermal coupling that implicitly captures differences in material response, spatial exposure, and relative position with respect to the thermal excitation.

Since \(\boldsymbol{\lambda}_i\) and \(\mathbf W_i\) are Gaussian-specific, primitives driven by the same scene-level process can exhibit different response speeds, amplitudes, and temporal trends, enabling spatially nonuniform temperature evolution. The dimension \(M\) controls the capacity of the shared latent process to represent multiple scene-level thermal modes, while \(\mathbf W_i\) determines how these modes contribute to individual Gaussians.

Since the base temperature $B_i$ is time-invariant, the temporal variation
of the Gaussian temperature is fully determined by the dynamic residual. Combining Eq.~\eqref{eq:local_thermal_state} with
Eq.~\eqref{eq:thermal_dynamic_residual}, the rate of temperature change with respect to the normalized temporal coordinate $\tau$ becomes

\begin{equation}
   \frac{\mathrm{d}\Delta T_i}{\mathrm{d}\tau}
    =
    -
    \mathbf{1}^{\mathsf T}
    \left(
    \boldsymbol{\lambda}_i
    \odot
    \mathbf{z}_i(\tau)
    \right)
    +
    \mathbf{1}^{\mathsf T}
    \mathbf{W}_i
    \mathbf{q}(\tau).
    \label{eq:gaussian_temperature_rate}
\end{equation}

Equation~\eqref{eq:gaussian_temperature_rate} shows that the Gaussian temperature variation is jointly determined by its local relaxation and the time-varying scene-level driving. Their balance determines both the magnitude and the sign of the temperature change.  Consequently, the same dynamics can represent increasing, decreasing, and non-monotonic temperature trajectories without prescribing their temporal forms in advance.

\textbf{Continuous rollout and future prediction.} All thermal dynamics above are formulated in the normalized temporal coordinate $\tau$ defined in Eq.~\eqref{eq:thermal_normalized_time}. The reference timestamp \(t_r\) is selected within the observed training interval, typically near its temporal midpoint. Accordingly, \(\tau(t_r)=0\), and the scene-level and Gaussian-local reference states are defined as

\begin{equation}
    \mathbf{h}_r=\mathbf{h}(0),
    \qquad
    \mathbf{z}_i^r=\mathbf{z}_i(0).
    \label{eq:thermal_reference_state}
\end{equation}
These reference states serve as learnable boundary conditions of the thermal dynamics. Note that $B_i$ is independent of the choice of $t_r$; the Gaussian temperature at the reference time is
\begin{equation}
    T_i(t_r)
    =
    B_i
    +
    \mathbf{1}^{\mathsf T}\mathbf{z}_i^r .
    \label{eq:temperature_at_reference}
\end{equation}

Starting from the reference state, the scene-level driver and Gaussian-local thermal state at any query time are obtained by integrating their corresponding differential equations:
\begin{equation}
\begin{aligned}
    \mathbf{h}(\tau)
    &=
    \mathbf{h}_r
    +
    \int_{0}^{\tau}
    f_{\psi}\!\left(\mathbf{h}(u)\right)\,
    \mathrm{d}u,\\
    \mathbf{z}_i(\tau)
    &=
    \mathbf{z}_i^r
    +
    \int_{0}^{\tau}
    \left[
    -
    \boldsymbol{\lambda}_i
    \odot
    \mathbf{z}_i(u)
    +
    \mathbf{W}_i
    \mathbf{q}(u)
    \right]
    \mathrm{d}u .
\end{aligned}
\label{eq:thermal_state_integral}
\end{equation}
When $\tau<0$, the states are integrated backward from $t_r$; when $\tau>0$, they are integrated forward. Therefore, observations on both sides of the reference time jointly constrain the same thermal dynamics. For timestamps beyond the observed training interval, the same learned dynamics are continued forward to enable future thermal prediction.

The coupled differential equations are numerically solved using a second-order Runge--Kutta (RK2) scheme. For one integration step from $\tau_n$ to
$\tau_{n+1}=\tau_n+\Delta\tau$, the derivatives at the current state are first evaluated as
\begin{equation}
\begin{aligned}
    \mathbf{k}^{h}_{1}
    &=
    f_{\psi}(\mathbf{h}_n),\\
    \mathbf{k}^{z}_{1,i}
    &=
    f_i(\mathbf{z}_{i,n},\mathbf{h}_n),
\end{aligned}
\label{eq:thermal_rk2_k1}
\end{equation}
where
\begin{equation}
    f_i(\mathbf{z}_i,\mathbf{h})
    =
    -
    \boldsymbol{\lambda}_i\odot\mathbf{z}_i
    +
    \mathbf{W}_i\tanh(\mathbf{h})
\end{equation}
is the local thermal derivative defined by
Eq.~\eqref{eq:local_thermal_state}. The derivatives are then evaluated at the midpoint:
\begin{equation}
\begin{aligned}
    \mathbf{k}^{h}_{2}
    &=
    f_{\psi}
    \left(
    \mathbf{h}_n+
    \frac{\Delta\tau}{2}\mathbf{k}^{h}_{1}
    \right),\\
    \mathbf{k}^{z}_{2,i}
    &=
    f_i
    \left(
    \mathbf{z}_{i,n}
    +
    \frac{\Delta\tau}{2}\mathbf{k}^{z}_{1,i},
    \mathbf{h}_n+
    \frac{\Delta\tau}{2}\mathbf{k}^{h}_{1}
    \right),
\end{aligned}
\label{eq:thermal_rk2_k2}
\end{equation}
and the states are updated by
\begin{equation}
\begin{aligned}
    \mathbf{h}_{n+1}
    &=
    \mathbf{h}_{n}
    +
    \Delta\tau\,\mathbf{k}^{h}_{2},\\
    \mathbf{z}_{i,n+1}
    &=
    \mathbf{z}_{i,n}
    +
    \Delta\tau\,\mathbf{k}^{z}_{2,i}.
\end{aligned}
\label{eq:thermal_rk2_update}
\end{equation}
The rollout is fully differentiable, so thermal observations at different timestamps can jointly optimize the reference states, scene-level dynamics, and Gaussian-specific responses.

\subsection{Thermal Parameterization and Inference Pipeline}
\label{sec:thermal_representation_optimization}

Having defined the Gaussian-level thermal dynamics, this subsection summarizes the learnable thermal parameters and the inference procedure for obtaining \(T_i(t)\) at a queried timestamp. The parameters are divided into Gaussian-specific variables and shared scene-level dynamics. For the $i$-th Gaussian, we define
\begin{equation}
    \Theta_i^{T}
    =
    \left\{
    B_i,\,
    \mathbf{z}_i^r,\,
   \hat{\boldsymbol{\lambda}}_i\,
    \mathbf{W}_i
    \right\},
    \label{eq:thermal_learnable_parameters_gaussian}
\end{equation}
where \(B_i\) is the base temperature, \(\mathbf z_i^r\) is the local reference state, \(\hat{\boldsymbol{\lambda}}_i\) is the unconstrained parameter from which the positive relaxation rates \(\boldsymbol{\lambda}_i\) are obtained, and \(\mathbf W_i\) determines the Gaussian-specific response to the shared thermal driver.

The scene-level dynamics are parameterized by

\begin{equation}
    \Theta_{\mathrm{scene}}^{T}
    =
    \left\{
    \mathbf{h}_r,\,
    \psi,\,
    \gamma_h
    \right\},
    \label{eq:thermal_learnable_parameters_scene}
\end{equation}
where \(\mathbf h_r\) is the reference scene-level state, \(\psi\) parameterizes the neural dynamics \(G_\psi\), and \(\gamma_h\) controls its temporal evolution rate.

For a query timestamp \(t\), the corresponding normalized coordinate \(\tau(t)\) is first computed using Eq.~\eqref{eq:thermal_normalized_time}, after which the scene-level and Gaussian-specific states are propagated from their reference states through the coupled thermal dynamics:

\begin{equation}
\begin{aligned}
&\Theta_{\mathrm{scene}}^{T}
\xrightarrow[\;u:0\rightarrow\tau(t)\;]{\mathrm{RK2\ rollout}}
\mathbf{h}(u)
\xrightarrow{\tanh}
\mathbf{q}(u),
\\
&\left(
\mathbf{z}_i^r,\,
\hat{\boldsymbol{\lambda}}_i,\,
\mathbf{W}_i,\,
\mathbf{q}(\cdot)
\right)
\xrightarrow[\;u:0\rightarrow\tau(t)\;]{\mathrm{RK2\ local\ rollout}}
\mathbf{z}_i(\tau(t)),
\\
&\left(
B_i,\,
\mathbf{z}_i(\tau(t))
\right)
\longrightarrow
T_i(t)
=
B_i+\mathbf{1}^{\mathsf T}\mathbf{z}_i(\tau(t)).
\end{aligned}
\label{eq:thermal_inference_pipeline}
\end{equation}

The shared scene-level dynamics capture the common temporal process, while the Gaussian-specific parameters convert it into spatially varying local thermal responses. Together with \(B_i\), they determine the queried Gaussian temperature \(T_i(t)\). The resulting temperatures are rendered through differentiable Gaussian rasterization, and the reconstruction loss is back-propagated through both the renderer and the temporal rollout to jointly optimize the scene-level and Gaussian-specific parameters.

\subsection{Integration with 3DGS and 4DGS Frameworks}
\label{sec:integration_3dgs_4dgs}

Within the proposed framework, the Gaussian backbone and GTSM play complementary roles. The backbone determines the spatial state of each Gaussian primitive, whereas GTSM provides its attached time-varying thermal state $T_i(t)$. Importantly, the geometric variables do not directly enter the GTSM state-evolution equations. Geometric motion changes the spatial support and projection of a Gaussian, while its thermal evolution is governed independently by the learned GTSM dynamics.

For a geometrically static scene, the Gaussian geometry remains unchanged over
time,
\begin{equation}
    \mathbf{x}_i(t)=\mathbf{x}_i,\qquad
    \mathbf{s}_i(t)=\mathbf{s}_i,\qquad
    \mathbf{r}_i(t)=\mathbf{r}_i,
\end{equation}
while $T_i(t)$ evolves continuously according to GTSM. For a dynamic scene, the
geometry is provided by the adopted dynamic Gaussian backbone,
\begin{equation}
    \left\{
    \mathbf{x}_i(t),
    \mathbf{s}_i(t),
    \mathbf{r}_i(t)
    \right\}
    =
    \mathcal{F}_{\mathrm{geo}}
    \left(
    \mathcal{G}_i^{0}, t
    \right),
\end{equation}
whereas the thermal state of the corresponding Gaussian is queried independently as
\begin{equation}
    T_i(t)
    =
    \mathcal{F}_{\mathrm{thermal}}
    \left(
    \Theta_i^{T},
    \Theta_{\mathrm{scene}}^{T},
    t
    \right).
\end{equation}

This separation allows the same GTSM thermal dynamics to be coupled with either a static 3DGS representation or a time-varying 4DGS representation, depending on the geometric characteristics of the observed scene.

\textbf{Integration with the 3DGS framework.}
For the static-geometry setting, the data loading and reconstruction pipeline of ETGS is adopted, with its original temporal temperature model replaced by GTSM. The input consists of infrared images acquired from different viewpoints and
timestamps together with the corresponding camera poses. Since the target geometry is approximately static, the same Gaussian geometry is shared across all timestamps, while the attached thermal state $T_i(t)$ evolves according to GTSM.

For each observed timestamp $t_k$, the temperature of every Gaussian is first queried from the learned thermal dynamics. Following ETGS~\cite{wangetgs}, the predicted temperatures are linearly mapped to normalized infrared grayscale values using the scene-specific calibrated temperature bounds provided by RHD
and are subsequently rendered using the standard differentiable Gaussian rasterizer. Since no RGB supervision is available in this setting, the reconstructed infrared image provides the main observation constraint. The image-level reconstruction objective is
\begin{equation}
    \mathcal{L}_{\mathrm{ir}}
    =
    (1-\lambda_{\mathrm{ssim}})
    \left\|
    \hat{\mathbf{I}}^{\mathrm{ir}}_k
    -
    \mathbf{I}_{k}^{\mathrm{ir,gt}}
    \right\|_1
    +
    \lambda_{\mathrm{ssim}}
    \left[
    1-
    \operatorname{SSIM}
    \left(
    \hat{\mathbf{I}}^{\mathrm{ir}}_k,
    \mathbf{I}_{k}^{\mathrm{ir,gt}}
    \right)
    \right],
    \label{eq:ir_image_loss}
\end{equation}
where $\hat{\mathbf{I}}^{\mathrm{ir}}_k$ and
$\mathbf{I}_{k}^{\mathrm{ir,gt}}$ denote the rendered and observed infrared
images, respectively. The reconstruction loss is back-propagated through the
Gaussian renderer and GTSM rollout, thereby jointly optimizing the static
Gaussian geometry and thermal dynamics from multi-view infrared observations.

\textbf{Integration with the 4DGS framework.}
For targets with time-varying geometry, $\mathcal{F}_{\mathrm{geo}}$ is instantiated using TRACE~\cite{li2025trace}. TRACE represents the dynamic scene by deforming a canonical Gaussian representation over time, while GTSM is attached to the same Gaussian primitives and independently provides their thermal states $T_i(t)$. Consequently, geometric deformation determines where each primitive is located and rendered, whereas its temperature evolution is governed by the GTSM dynamics introduced in Sec.~\ref{sec:thermal_state_space}.

Optimization is performed in two stages. In the first stage, the dynamic Gaussian geometry and deformation field are optimized from RGB observations following the original TRACE training strategy. The rendered RGB images are
supervised using an L1--SSIM objective analogous to
Eq.~\eqref{eq:ir_image_loss}, denoted as $\mathcal{L}_{\mathrm{rgb}}$.
This stage provides a stable time-varying spatial representation of the target.

In the second stage, GTSM is attached to the reconstructed dynamic Gaussians and
optimized using the corresponding two-dimensional temperature observations.
For each timestamp $t_k$, GTSM predicts the Gaussian-level temperatures
$\{T_i(t_k)\}_{i=1}^{N}$, which are rendered into a temperature map
$\hat{\mathbf{T}}_k$. The thermal state-space parameters are optimized using
the temperature-map MAE:
\begin{equation}
    \mathcal{L}_{T}
    =
    \frac{1}{|\Omega_k|}
    \sum_{\mathbf{p}\in\Omega_k}
    \left|
    \hat{\mathbf{T}}_k(\mathbf{p})
    -
    \mathbf{T}_{k}^{\mathrm{gt}}(\mathbf{p})
    \right|,
    \label{eq:temperature_mae_loss}
\end{equation}
where $\Omega_k$ denotes the valid target region of the temperature map.
Accordingly, the RGB observations establish the time-varying geometric support,
while the temperature observations recover the thermal evolution on the same
moving Gaussian representation.

\section{Experiments}
\label{sec4}

\subsection{Dataset Details}
\label{sec:dataset_details}

GTSM is evaluated on two complementary types of data: a real infrared dataset containing actual thermal evolution and a synthetic dynamic thermal dataset with controllable geometric and thermal variations. The RHD dataset represents the static-geometry setting, where the target geometry remains approximately unchanged while the camera moves and the thermal state evolves over time. The synthetic dataset further introduces time-varying target geometry and is used to evaluate GTSM when both geometry and temperature evolve simultaneously within a 4DGS framework.

\textbf{Real infrared dataset.} The real-data evaluation adopts the Rapid Heat Dynamics (RHD) dataset introduced by ETGS~\cite{wangetgs}. RHD contains real RGB--IR image sequences of multiple objects undergoing practical thermal processes, including warming, cooling, heating, and heat transfer. The sequences are acquired under continuous camera motion with millisecond-level timestamps, providing temporal infrared observations of real thermal evolution. The thermal observations are radiometrically calibrated, and the dataset provides the environmental temperature together with scene-specific temperature bounds corresponding to the infrared grayscale measurements.

The infrared sequences are mainly used to evaluate time-varying thermal-field reconstruction from moving-camera observations and future thermal-state prediction beyond the observed interval. Since the targets in RHD are approximately geometrically static, this dataset corresponds to the 3DGS-based setting described in Sec.~\ref{sec:integration_3dgs_4dgs}.

\textbf{Synthetic dynamic thermal dataset.} Although RHD provides real thermal observations, datasets simultaneously containing multi-view images, dynamic target motion, and accurate time-varying temperature ground truth remain limited. To evaluate GTSM under dynamic geometry, a thermal-augmented dynamic dataset is therefore constructed based on D-NeRF~\cite{pumarola2020d}. The original D-NeRF dataset provides multi-view RGB images, camera poses, dynamic target motion, and normalized timestamps. Its camera trajectories and target motion are retained, while corresponding temperature maps are synthesized by assigning simulated component-level temperatures to the foreground target regions.

Each target is segmented into semantic components, such as the seat and frame
of a folding chair or the body and solar panels of a satellite. Views
corresponding to the same timestamp share identical component-level
temperatures, ensuring cross-view consistency of the underlying thermal state.
The component temperatures are generated using a simplified lumped thermal ODE
model. For component $r$, the energy-balance equation is written as

\begin{align}
C_r \frac{\mathrm{d}T_r}{\mathrm{d}t}
&=
Q_r(t)
+
\sum_q G_{rq}\left(T_q-T_r\right)
\nonumber\\
&\quad
-
\chi_r^{\mathrm{conv}} h_r A_r^{\mathrm{rad}}
\left(T_r-T_{\mathrm{amb}}\right)
\nonumber\\
&\quad
-
\varepsilon_r \sigma A_r^{\mathrm{rad}}
\left(
T_{r,K}^{4}-T_{\mathrm{rad},K}^{4}
\right).
\label{eq:component_thermal_ode}
\end{align}

Here, $T_r$ is the component temperature in Celsius, and
$T_{r,K}=T_r+273.15$ is its corresponding temperature in Kelvin.
$C_r$ denotes the effective heat capacity, $Q_r(t)$ represents the heat input,
including absorbed solar heating and internal heat dissipation, and $G_{rq}$
denotes the thermal conductance between components. The coefficient $h_r$ is
the convective heat-transfer coefficient, $A_r^{\mathrm{rad}}$ is the effective
surface area used for heat exchange, $\varepsilon_r$ is the emissivity, and
$\sigma$ is the Stefan--Boltzmann constant. $T_{\mathrm{amb}}$ denotes the
ambient air temperature used for convection, whereas $T_{\mathrm{rad},K}$
denotes the effective radiative-background temperature in Kelvin. The switch
$\chi_r^{\mathrm{conv}}$ controls whether convection is enabled. For the satellite simulations, convection is disabled to approximate a vacuum environment, i.e., \(\chi_r^{\mathrm{conv}}=0\), and heat exchange with the surroundings is dominated by radiation and inter-component conduction.

Five representative thermal cases are generated by varying the initial temperature, ambient condition, heat-source mode, and heat-input strength: natural warming, natural cooling, constant heating, ramp heating, and pulse heating--cooling. Natural warming and cooling represent environmentally driven thermal transitions without an explicitly prescribed active heating profile. Constant heating uses a continuous heat input. Ramp heating gradually increases the heat-source strength. Pulse heating--cooling applies heat in the early stage and then removes it, producing a non-monotonic temperature curve. These cases provide different temperature ranges and temporal trends for evaluating thermal reconstruction and future-time prediction.

\begin{figure}[t]
    \centering
    \includegraphics[width=\columnwidth]{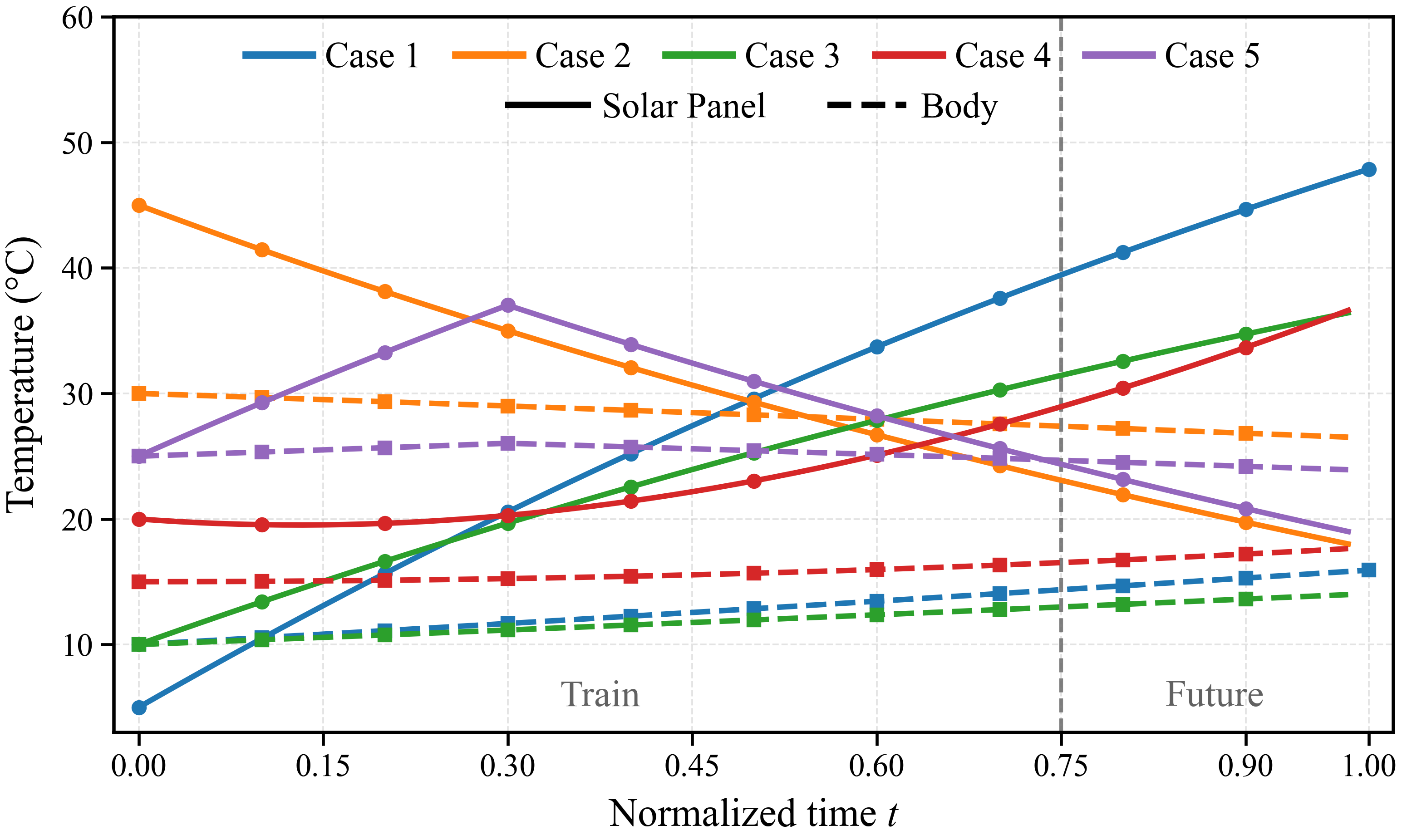}
    \caption{ Simulated component-level temperature curves of the satellite target under five ODE-based thermal cases. Solid and dashed curves denote different target components. The vertical dashed line at $t=0.75$ separates the observed interval from the future-time prediction interval.
    }
    \label{fig:ode_temperature_curves}
\end{figure}

The simulated component temperatures are assigned to the corresponding segmented regions to form per-frame two-dimensional temperature maps. This construction separates the three controlled factors: the D-NeRF camera poses determine viewpoint variation, the original sequence provides time-varying geometry, and the component-level thermal ODE governs temperature evolution. Fig.~\ref{fig:ode_temperature_curves} shows the simulated component-level temperature curves under the five ODE-based thermal cases. 

Although the simulation does not reproduce detailed spatial heat diffusion or material-interface effects, it provides controllable component-level temperature ground truth and therefore enables quantitative evaluation of spatio-temporal thermal reconstruction and future prediction under dynamic geometry.

\textbf{Dataset split and evaluation protocol.} For the RHD dataset, the target geometry is approximately static, whereas the camera moves continuously and the temperature evolves over time. Two evaluation settings are considered. For within-interval reconstruction, one out of every eight consecutive frames is held out for testing, with the remaining frames used for training. This evaluates interpolation at unseen viewpoints and timestamps within the observed temporal range. For temporal extrapolation, the first 80\% of each sequence is used for training and the remaining 20\% for testing, requiring prediction at future timestamps and corresponding unseen viewpoints.

For the synthetic dynamic thermal dataset, time is normalized to \([0,1]\). All RGB and temperature observations within \(t\in[0,0.75]\) are used for training, while \(t\in(0.75,1]\) is completely reserved for future-time testing. The validation set contains novel viewpoints within the observed interval and evaluates within-interval spatial generalization. The test set contains both unseen viewpoints and future timestamps, and therefore evaluates joint future geometry and thermal-state prediction.

Together, the two protocols evaluate GTSM across complementary settings: real thermal evolution with approximately static geometry and synthetic thermal evolution with time-varying geometry, covering within-interval reconstruction, novel-view rendering, and future-time thermal prediction.

\subsection{Implementation Details}
\label{sec:implementation_details}

For the synthetic dynamic thermal dataset, the training follows a geometry-first strategy. Each scene is trained for $45{,}000$ iterations in total. In the first 30,000 iterations, we train the dynamic Gaussian geometry using only the RGB observations within the training interval \(t\in[0,0.75]\), following the original TRACE training strategy.  After the dynamic geometry is obtained, we activate GTSM and train the Gaussian-level thermal dynamics for another $15{,}000$ iterations using the paired temperature maps. This design decouples dynamic geometry learning from thermal state learning, allowing the RGB sequence to provide reliable time-dependent spatial support for subsequent temperature-field reconstruction. No RGB or temperature observations from the future test interval are used during either training stage. 

For the real RHD infrared dataset, we train each scene for $30{,}000$ iterations, following the main optimization settings of the 3DGS/ETGS reconstruction pipeline. The proposed GTSM replaces the original thermal evolution model, and the rendered infrared image is supervised by the image-level reconstruction loss. Thus, both the Gaussian geometry and the Gaussian-level thermal state-space parameters are updated from the infrared image sequence.

In both settings, Gaussian densification follows the standard 3DGS strategy and is enabled from iteration $500$ to $15{,}000$. During Gaussian cloning or splitting, the associated GTSM parameters \(B_i\), \(z_i^r\), \(\hat{\lambda}_i\), and \(\rm{W}_i\) are inherited from the parent Gaussian, while pruning removes the corresponding thermal states together with the Gaussian primitive. The image-level reconstruction loss adopts the L1-SSIM form, where $\lambda_{\mathrm{ssim}}=0.2$. 

For GTSM, we set the local thermal state dimension to $D=3$ and the scene-level driver dimension to $M=4$. The hidden dimension of the scene-level driver MLP is set to $64$. The thermal reference state is placed at the midpoint of the training time range by default. The coupled scene-level and Gaussian-level thermal states are integrated using RK2 rollout, with $24$ integration steps per unit normalized time and a maximum of $128$ steps. The Gaussian base temperature, local reference state, relaxation rates, coupling matrix, and scene-level driver parameters are all optimized during training.

\subsection{Results on the Real RHD Dataset}
\label{sec:rhd_results}

We first evaluate GTSM on the real RHD dataset, which contains infrared sequences of different objects undergoing practical thermal processes. Following the protocol in Sec.~\ref{sec:dataset_details}, we consider two settings: within-interval novel-view synthesis and temporal extrapolation. The former evaluates thermal-field reconstruction from moving-camera infrared observations, while the latter evaluates future thermal-state prediction. PSNR, SSIM, and LPIPS are used as image-level metrics.

\begin{table*}[t]
\centering
\caption{
Quantitative comparison on the real RHD dataset. The upper part reports within-interval novel-view/temporal interpolation results, while the lower part reports temporal extrapolation results. Higher PSNR and SSIM and lower LPIPS indicate better performance.
}
\label{tab:rhd_quantitative_results}

\renewcommand{\arraystretch}{1.10}
\setlength{\tabcolsep}{1.6pt}
\footnotesize

\begin{adjustbox}{max width=\textwidth}
\begin{tabular}{c c *{10}{c} c}
\toprule

\multirow{2}{*}{\textbf{Metric}}
& \multirow{2}{*}{\textbf{Method}}
& \textbf{Cooling}
& \textbf{Warming}
& \textbf{Cooling}
& \textbf{Cooling}
& \textbf{Cooling}
& \textbf{Heat}
& \textbf{Warming}
& \textbf{Heating}
& \textbf{Warming}
& \textbf{Warming}
& \multirow{2}{*}{\textbf{Avg}} \\

&
& \scriptsize \textbf{Checkboard}
& \scriptsize \textbf{Bottles}
& \scriptsize \textbf{Dumbbells}
& \scriptsize \textbf{Bench}
& \scriptsize \textbf{Ebike}
& \scriptsize \textbf{Transfer}
& \scriptsize \textbf{Peaches}
& \scriptsize \textbf{Workpiece}
& \scriptsize \textbf{Cups}
& \scriptsize \textbf{Workpieces}
& \\

\midrule
\multicolumn{13}{c}{\textbf{Within-interval Novel-view/Temporal Interpolation}} \\
\midrule

\multirow{6}{*}{PSNR $\uparrow$}
& 3DGS
& 36.27 & 32.37 & 29.78 & 29.74 & 34.66
& 39.47 & 30.48 & 34.47 & 24.43 & 29.94 & 32.16 \\

& Thermal3D-GS
& 39.47 & 33.61 & 32.89 & 32.11 & 38.77
& 39.86 & 34.12 & 34.93 & 30.96 & 30.06 & 34.68 \\

& 4D-GS
& 33.58 & 30.52 & 34.50 & 35.61 & 33.99
& 39.75 & 34.64 & 30.36 & 34.15 & 32.34 & 33.94 \\

& NTR-Gaussian
& 41.82 & 34.63 & 35.18 & 35.56 & 38.09
& 38.99 & 32.44 & 33.65 & 28.18 & 31.01 & 34.96 \\

& ETGS
& 43.14 & 40.25 & \textbf{36.05} & 40.98 & 41.46
& 40.03 & 42.85 & \textbf{37.22} & 38.21 & 42.16 & 40.23 \\

& Ours
& \textbf{44.86} & \textbf{43.19} & 35.06 & \textbf{42.38} & \textbf{41.70}
& \textbf{40.77} & \textbf{43.37} & 35.68 & \textbf{40.77} & \textbf{43.69}
& \textbf{41.15} \\

\midrule

\multirow{6}{*}{SSIM $\uparrow$}
& 3DGS
& 0.978 & 0.988 & 0.968 & 0.960 & 0.980
& 0.990 & 0.987 & 0.988 & 0.968 & 0.974 & 0.978 \\

& Thermal3D-GS
& 0.983 & 0.989 & 0.981 & 0.970 & 0.983
& 0.990 & 0.990 & 0.989 & 0.978 & 0.976 & 0.983 \\

& 4D-GS
& 0.963 & 0.979 & 0.976 & 0.967 & 0.962
& 0.986 & 0.984 & 0.960 & 0.975 & 0.970 & 0.972 \\

& NTR-Gaussian
& 0.986 & 0.987 & 0.980 & 0.974 & 0.974
& 0.987 & 0.984 & 0.982 & 0.974 & 0.978 & 0.981 \\

& ETGS
& 0.987 & 0.990 & \textbf{0.986} & 0.989 & \textbf{0.985}
& 0.990 & 0.995 & 0.986 & 0.987 & 0.993 & 0.989 \\

& Ours
& \textbf{0.990} & \textbf{0.993} & 0.985 & 0.989 & 0.984
& \textbf{0.992} & \textbf{0.996} & \textbf{0.988} & \textbf{0.987}
& \textbf{0.995} & \textbf{0.990} \\

\midrule

\multirow{6}{*}{LPIPS $\downarrow$}
& 3DGS
& 0.066 & 0.087 & 0.072 & 0.075 & 0.031
& 0.057 & 0.091 & 0.031 & 0.131 & 0.137 & 0.078 \\

& Thermal3D-GS
& 0.060 & 0.095 & 0.050 & 0.063 & 0.029
& 0.062 & 0.092 & \textbf{0.030} & 0.110 & 0.128 & 0.072 \\

& 4D-GS
& 0.101 & 0.075 & 0.047 & 0.076 & 0.065
& 0.056 & 0.064 & 0.073 & 0.105 & 0.099 & 0.076 \\

& NTR-Gaussian
& 0.072 & 0.092 & 0.052 & 0.092 & 0.098
& 0.092 & 0.104 & 0.055 & 0.133 & 0.102 & 0.089 \\

& ETGS
& 0.060 & 0.087 & 0.039 & 0.051 & 0.029
& 0.073 & \textbf{0.055} & 0.032 & 0.096 & \textbf{0.061} & 0.058 \\

& Ours
& \textbf{0.057} & \textbf{0.065} & 0.039 & \textbf{0.039} & \textbf{0.025}
& \textbf{0.061} & 0.073 & 0.036 & \textbf{0.089} & 0.068
& \textbf{0.055} \\

\midrule
\multicolumn{13}{c}{\textbf{Temporal Extrapolation}} \\
\midrule

\multirow{3}{*}{PSNR $\uparrow$}
& ETGS
& 39.18 & 32.90 & 28.27 & 22.88 & 30.37
& 29.45 & 22.20 & 27.87 & 39.15 & 22.58 & 29.49 \\

& MLP
& 46.43 & 37.56 & 33.32 & 37.10 & 40.76
& 37.66 & 36.67 & 33.82 & 38.65 & 39.11 & 38.11 \\

& Ours
& \textbf{47.91} & \textbf{42.10} & \textbf{34.43} & \textbf{40.82}
& \textbf{44.03} & \textbf{39.99} & \textbf{41.06} & \textbf{36.80}
& \textbf{41.11} & \textbf{41.50} & \textbf{40.98} \\

\midrule

\multirow{3}{*}{SSIM $\uparrow$}
& ETGS
& 0.989 & 0.977 & 0.958 & 0.705 & 0.933
& 0.991 & 0.983 & 0.971 & 0.985 & 0.985 & 0.948 \\

& MLP
& 0.990 & 0.990 & 0.975 & 0.985 & 0.989
& 0.992 & 0.993 & 0.985 & 0.989 & 0.991 & 0.988 \\

& Ours
& \textbf{0.992} & \textbf{0.994} & \textbf{0.984} & \textbf{0.990}
& \textbf{0.991} & \textbf{0.993} & \textbf{0.995} & \textbf{0.990}
& \textbf{0.991} & \textbf{0.992} & \textbf{0.991} \\

\midrule

\multirow{3}{*}{LPIPS $\downarrow$}
& ETGS
& 0.058 & 0.104 & 0.093 & 0.071 & 0.071
& 0.068 & 0.110 & 0.066 & 0.107 & 0.120 & 0.087 \\

& MLP
& 0.073 & 0.087 & 0.081 & 0.044 & 0.031
& 0.083 & 0.101 & 0.052 & 0.128 & 0.133 & 0.081 \\

& Ours
& \textbf{0.046} & \textbf{0.053} & \textbf{0.047} & \textbf{0.041}
& \textbf{0.023} & \textbf{0.060} & \textbf{0.079} & \textbf{0.040}
& \textbf{0.073} & \textbf{0.088} & \textbf{0.055} \\

\bottomrule
\end{tabular}
\end{adjustbox}
\end{table*}

Table~\ref{tab:rhd_quantitative_results} reports the quantitative comparison on the real RHD dataset. For within-interval novel-view/temporal interpolation, we compare against 3DGS, Thermal3D-GS, 4D-GS, NTR-Gaussian, and ETGS. For temporal extrapolation, methods that only reconstruct or interpolate thermal appearance within the observed interval are not included, since they do not explicitly provide a continuous thermal-evolution model for future-time prediction under this setting. We therefore compare GTSM with ETGS, which explicitly models continuous thermal evolution on RHD, and additionally introduce a direct MLP baseline to evaluate whether future temperature can be predicted using a generic time-conditioned neural representation.

For the MLP baseline, a shared network directly predicts the temperature of each Gaussian from its spatial position and normalized query time, without introducing explicit thermal states, relaxation dynamics, or temporal rollout. This provides a simple data-driven reference for evaluating the benefit of the proposed structured thermal dynamics. Within the observed interval, GTSM achieves the best average PSNR of \(41.15\) dB, compared with \(40.23\) dB for ETGS, while maintaining comparable SSIM and LPIPS. The performance gap becomes substantially larger in temporal extrapolation: GTSM achieves \(40.98\) dB PSNR, compared with \(38.11\) dB for the direct MLP and \(29.49\) dB for ETGS. GTSM also achieves the best average SSIM and LPIPS, indicating that the proposed dynamics improve not only pixel-wise fidelity but also structural and perceptual consistency at future timestamps.

The different extrapolation behaviors are consistent with the temporal representations used by the three methods. The direct MLP learns an instantaneous mapping from position and time to temperature and can accurately fit observations within the training interval, but it does not explicitly model how a thermal state evolves from one time to another. Consequently, predictions for \(t\) outside the training range rely on direct function extrapolation. ETGS introduces stronger physical structure through a first-order thermal response model with exponential relaxation and harmonic heat-source terms. However, its future evolution remains constrained by the predefined response form and by sinusoidal forcing bases estimated from the observed interval, which can limit its ability to represent thermal trends that deviate from these assumed temporal patterns. In contrast, GTSM learns a continuous latent thermal process together with Gaussian-specific relaxation and coupling dynamics, and future temperatures are obtained by propagating the learned states beyond the observed interval. The consistent improvement over both ETGS and the direct MLP therefore suggests that explicitly modeling thermal-state evolution provides a more effective inductive bias for future-time thermal prediction than either a prescribed closed-form trajectory or direct time-to-temperature regression.

\begin{figure*}[t]
    \centering
    \includegraphics[width=\textwidth]{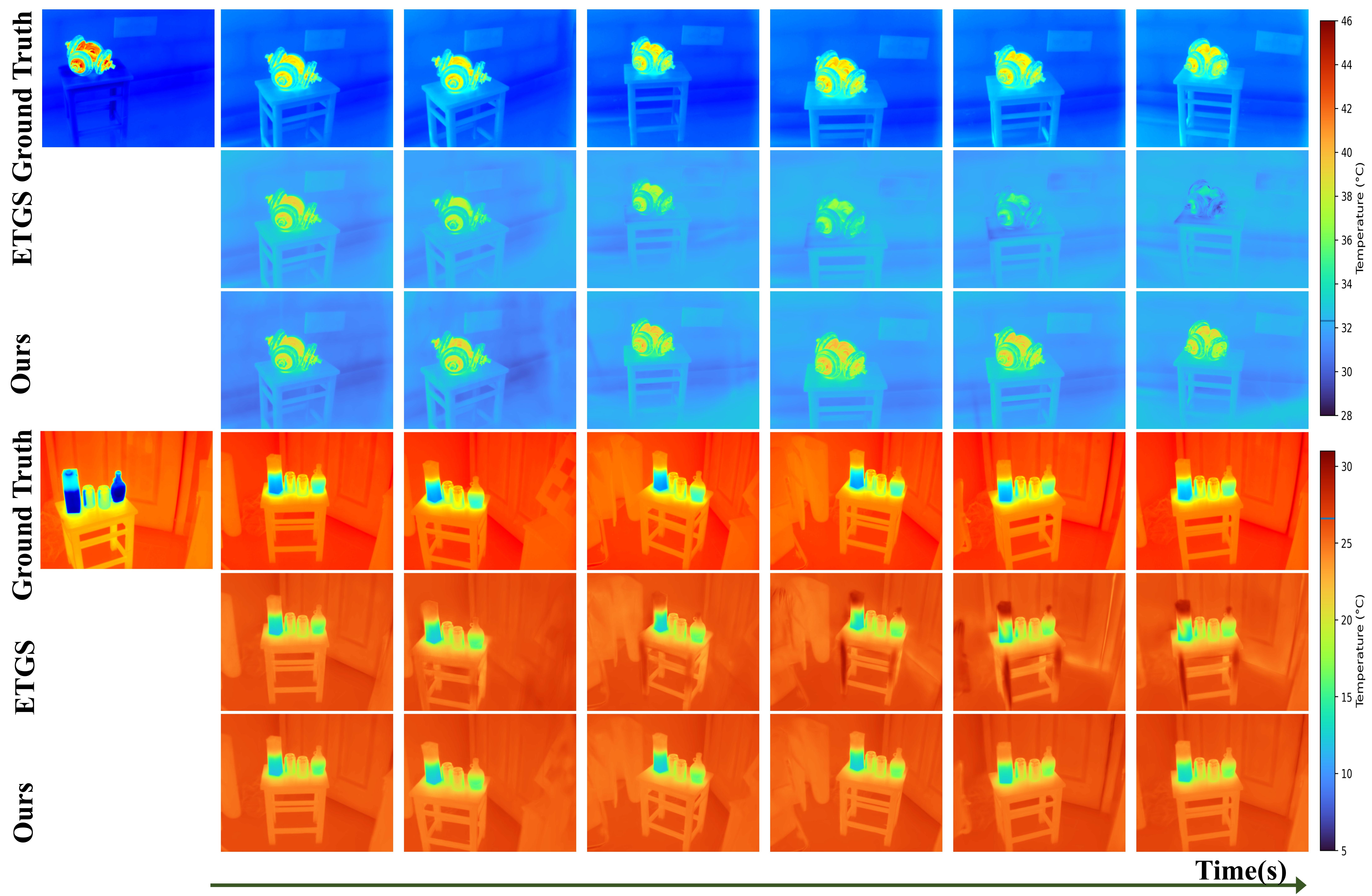}
    \caption{
    Qualitative comparison of temporal extrapolation on the RHD dataset for the Cooling Dumbbells and Warming Bottles scenes. The first ground-truth image shows the beginning of the training sequence, while the remaining columns show future-time test frames in chronological order.
    }
    \label{fig:rhd_temporal_extrapolation}
\end{figure*}

Fig.~\ref{fig:rhd_temporal_extrapolation} presents qualitative extrapolation results on the Cooling Dumbbells and Warming Bottles scenes. The first ground-truth image shows the beginning of the training sequence, while the remaining columns correspond to progressively farther future timestamps. Although both methods provide reasonable predictions near the observed interval, ETGS gradually deviates from the ground-truth thermal distribution as the extrapolation horizon increases. In contrast, GTSM better preserves the spatial temperature distribution and the corresponding cooling or warming trend over longer prediction horizons, consistent with the quantitative results.

\subsection{Results on the Synthetic Dynamic Thermal Dataset}
\label{sec:synthetic_dynamic_results}

We further evaluate GTSM within a 4DGS framework on the D-NeRF-based synthetic dynamic thermal dataset, where both target geometry and temperature evolve over time. Five component-level thermal processes are considered, including natural warming, natural cooling, constant heating, ramp heating, and pulse
heating--cooling. MAE and RMSE are used to evaluate temperature accuracy, while PSNR, SSIM, and LPIPS measure the image-level quality of the rendered RGB and pseudo-color temperature results. The validation set evaluates novel-view
reconstruction within the observed interval, whereas the test set evaluates future geometry and thermal-state prediction beyond it.

Table~\ref{tab:thermal_quantitative_results} reports the temperature accuracy.
To isolate thermal modeling from geometry reconstruction, all thermal cases of
the same target share the same dynamic geometry model, pre-trained for
$30{,}000$ iterations using the RGB sequence, followed by $15{,}000$ iterations
of thermal learning. Therefore, differences among the thermal cases mainly
reflect the ability of GTSM to model different temperature evolution processes.

A constant-temperature setting is first used as a reference, where the remaining temperature error mainly reflects geometry, projection, visibility, and rasterization effects. Across the five dynamic thermal cases, GTSM achieves low MAE and RMSE on the validation set, demonstrating accurate reconstruction of
component-dependent temperature evolution within the observed interval. The errors increase moderately on the test set because both geometry and thermal states must be extrapolated, but the prediction remains stable across different warming, cooling, heating, and non-monotonic processes. This indicates that GTSM is not restricted to a predefined thermal trajectory and can propagate learned local thermal dynamics to future timestamps under dynamic geometry. The RMSE is consistently higher than the MAE because a small number of pixels near moving object boundaries can incur relatively large temperature errors when the extrapolated geometry is slightly misaligned with the ground truth, whereas the temperature prediction over most object-interior regions remains accurate.

\begin{table*}[t]
\centering
\caption{
Quantitative evaluation under different targets, temperature ranges, and temporal evolution laws. Validation metrics measure novel-view temperature-field reconstruction within the observed interval ($t\leq0.75$), while test metrics measure future-time temperature prediction beyond the observed interval ($t>0.75$).
}
\label{tab:thermal_quantitative_results}
\renewcommand{\arraystretch}{1.15}
\begin{tabular}{c c c c c c c}
\toprule
\textbf{Target}
& \textbf{Schedule}
& \textbf{$T(0)\rightarrow T(0.75)\rightarrow T(1.0)$ $(^\circ\mathrm{C})$}
& \textbf{Val MAE}
& \textbf{Val RMSE}
& \textbf{Test MAE}
& \textbf{Test RMSE} \\
\midrule

\multirow{6}{*}{\makecell[c]{Foldingchair\\Seat--Frame}}

& Constant
& 15,20 $\rightarrow$ 15,20 $\rightarrow$ 15,20
& 0.0608 & 0.9748 & 0.2035 & 1.8513 \\

& Case1:natural warming
& 10,5 $\rightarrow$ 12.89,17.26 $\rightarrow$ 13.70,19.20
& 0.0438 & 0.6806 & 0.1549 & 1.4938 \\

& Case2:natural cooling
& 35,40 $\rightarrow$ 31.03,24.56 $\rightarrow$ 29.93,22.19
& 0.1126 & 1.8314 & 0.2522 & 2.4806 \\

& Case3:constant heating
& 15,15 $\rightarrow$ 19.90,17.54 $\rightarrow$ 21.21,17.97
& 0.0584 & 0.9683 & 0.1747 & 1.7681 \\

& Case4:ramp heating
& 15,15 $\rightarrow$ 18.95,20.48 $\rightarrow$ 21.65,23.66
& 0.0572 & 0.9635 & 0.1981 & 1.9796 \\

& Case5:heating-cooling
& 20,20 $\rightarrow$ 21.05,16.97 $\rightarrow$ 20.32,15.27
& 0.0779 & 1.1729 & 0.2208 & 1.7418 \\

\midrule

\multirow{6}{*}{\makecell[c]{Satellite\\Solar Panel--Body}}
& Constant
& 20,25 $\rightarrow$ 20,25 $\rightarrow$ 20,25
& 0.0671 & 1.0675 & 0.0953 & 1.2842 \\

& Case1
& 5,10 $\rightarrow$ 39.44,14.36 $\rightarrow$ 47.87,15.92
& 0.0757 & 1.0714 & 0.1278 & 1.6433 \\

& Case2
& 50,45 $\rightarrow$ 28.08, 41.36 $\rightarrow$ 22.70,40.11
& 0.0979 & 1.5045 & 0.1186 & 1.5766 \\

& Case3
& 10,10 $\rightarrow$ 31.44,12.99 $\rightarrow$ 36.75,14.05
& 0.0576 & 0.8559 & 0.0954 & 1.2417 \\

& Case4
& 20,15 $\rightarrow$ 28.95,16.52 $\rightarrow$ 37.24,17.73
& 0.0590 & 0.9217 & 0.1220 & 1.3028 \\

& Case5
& 25,25 $\rightarrow$ 24.36,24.67 $\rightarrow$ 18.64, 23.87
& 0.0948 & 1.4009 & 0.1153 & 1.4273 \\

\bottomrule
\end{tabular}
\end{table*}

\begin{table}[t]
\centering
\caption{
Image-level evaluation on the synthetic dynamic thermal dataset. TRACE and RGB row reports geometric rendering quality, while Cases 1--5 report pseudo-color temperature reconstruction and prediction results.}
\label{tab:image_level_metrics_val_test}
\renewcommand{\arraystretch}{1.08}
\setlength{\tabcolsep}{1.5pt}
\scriptsize
\resizebox{\columnwidth}{!}{

\begin{tabular}{c c c c c c c c}
\toprule
\multirow{2}{*}{\textbf{Target}}
& \multirow{2}{*}{\textbf{Setting}}
& \multicolumn{3}{c}{\textbf{Val-Interpolation}}
& \multicolumn{3}{c}{\textbf{Test-Extrapolation}} \\
\cmidrule(lr){3-5}
\cmidrule(lr){6-8}
&
& \textbf{PSNR} $\uparrow$
& \textbf{SSIM} $\uparrow$
& \textbf{LPIPS} $\downarrow$
& \textbf{PSNR} $\uparrow$
& \textbf{SSIM} $\uparrow$
& \textbf{LPIPS} $\downarrow$ \\
\midrule

\multirow{7}{*}{Foldingchair}
& TRACE
& 40.03 & 0.996 & 0.005
& 29.59 & 0.982 & 0.010 \\

& Ours RGB
& 40.20 & 0.996 & 0.005
& 30.47 & 0.980 & 0.009 \\

\cmidrule(lr){2-8}

& Case 1
& 41.58 & 0.991 & 0.017
& 33.99 & 0.973 & 0.024 \\

& Case 2
& 32.70 & 0.985 & 0.022
& 29.13 & 0.969 & 0.026 \\

& Case 3
& 38.24 & 0.998 & 0.021
& 32.20 & 0.971 & 0.026 \\

& Case 4
& 38.36 & 0.989 & 0.022
& 31.41 & 0.970 & 0.027 \\

& Case 5
& 36.53 & 0.987 & 0.022
& 32.20 & 0.972 & 0.025 \\

\midrule

\multirow{7}{*}{Satellite}

& TRACE
&36.94 & 0.993 & 0.007
& 32.09 & 0.988 & 0.007\\

& Ours RGB
& 38.29 & 0.994 & 0.006
& 33.87 & 0.988 & 0.007 \\

\cmidrule(lr){2-8}

& Case 1
& 37.71 & 0.990 & 0.014
& 32.87 & 0.985 & 0.020 \\

& Case 2
& 34.17 & 0.990 & 0.014
& 33.02 & 0.986 & 0.018\\

& Case 3
& 39.38 & 0.991 & 0.013
& 35.51 & 0.986 & 0.018 \\

& Case 4
& 38.50 & 0.991 & 0.013
& 34.97 & 0.986 & 0.018 \\

& Case 5
& 34.69 & 0.989 & 0.014
& 33.83 & 0.987 & 0.017 \\

\bottomrule
\end{tabular}
}
\end{table}

Table~\ref{tab:image_level_metrics_val_test} further reports image-level rendering quality. The RGB results of the original TRACE framework and the geometry model used in our method remain comparable on both validation and test sets, showing that the thermal modeling stage does not noticeably degrade the
underlying dynamic geometry or RGB rendering quality. The pseudo-color temperature results also maintain high reconstruction quality within the observed interval and remain stable during future-time extrapolation. Overall, GTSM preserves the geometric rendering capability of the 4DGS backbone while enabling consistent thermal reconstruction and future thermal-state prediction for dynamic targets.

\begin{figure}[ht]
    \centering
    \includegraphics[width=\columnwidth]{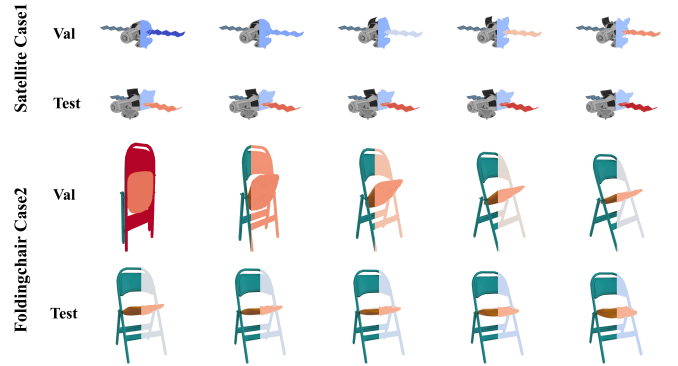}
    \caption{
    Qualitative results of GTSM integrated with the 4DGS framework on the
    Satellite and Foldingchair scenes. Each visualization combines the rendered
    RGB appearance and pseudo-color temperature map. The validation results show
    novel-view reconstruction within the observed temporal interval, while the
    test results simultaneously evaluate novel-view synthesis and future thermal
    prediction beyond the observed interval.
    }
    \label{fig:dynamic_thermal_visualization}
\end{figure}

Fig.~\ref{fig:dynamic_thermal_visualization} shows qualitative results of GTSM integrated with 4DGS for dynamic targets. The model achieves consistent geometry and temperature reconstruction within the observed interval and further supports novel-view synthesis and future thermal-state prediction beyond the observed timestamps.

\subsection{Ablation and Additional Analysis}

\begin{table}[t]
    \centering
    \caption{Ablation study of the major components in GTSM for temporal interpolation and extrapolation.}
    \label{tab:gtsm_component_ablation}
    \scriptsize
    \setlength{\tabcolsep}{2.0pt}
    \begin{adjustbox}{max width=\columnwidth}
    \begin{tabular}{lcccccc}
        \toprule
        & \multicolumn{3}{c}{Interpolation}
        & \multicolumn{3}{c}{Extrapolation} \\
        \cmidrule(lr){2-4}
        \cmidrule(lr){5-7}
        Method
        & PSNR$\uparrow$ & SSIM$\uparrow$ & LPIPS$\downarrow$
        & PSNR$\uparrow$ & SSIM$\uparrow$ & LPIPS$\downarrow$ \\
        \midrule
        Full model
        & \textbf{41.15} & 0.990 & \textbf{0.055}
        & \textbf{40.98} & 0.991 & \textbf{0.055} \\

        w/o learnable $B_i$
        & 40.96 & 0.991 & 0.058
        & 39.92 & 0.991 & 0.058 \\

        w/o scene driver
        & 40.52 & 0.990 & 0.060
        & 40.65 & \textbf{0.992} & 0.059 \\

        w/o relaxation
        & 40.61 & 0.991 & 0.059
        & 38.37 & 0.990 & 0.060 \\
        \bottomrule
    \end{tabular}
    \end{adjustbox}
\end{table}

Table~\ref{tab:gtsm_component_ablation} reports component-wise ablations of the learnable thermal base, scene-level thermal driver, and local relaxation term. Results on both interpolation and extrapolation are included to evaluate their roles in observed-interval reconstruction and future-time prediction.

\textbf{Effect of the learnable thermal base.}
Removing the learnable base temperature $B_i$ forces the model to explain most of the observed temperature variation through the dynamic residual $\mathbf{z}_i(\tau)$. Although the interpolation performance is only slightly affected, the extrapolation PSNR decreases from 40.98~dB to 39.92~dB. This indicates that a Gaussian-specific persistent thermal level provides a more stable reference around which the dynamic thermal state can evolve, particularly when predicting beyond the observed interval.

\textbf{Effect of the scene-level thermal driver.}
Setting the Gaussian--driver coupling $\mathbf{W}_i$ to zero removes the scene-level forcing term, such that the local state evolves only according to its Gaussian-specific relaxation dynamics. This variant produces lower PSNR and higher LPIPS for both interpolation and extrapolation, indicating that local relaxation alone cannot sufficiently explain the common temporal thermal trend shared across the target. The scene-level latent driver is therefore important for capturing global heating or cooling behavior from the observed sequence.

\textbf{Effect of local thermal relaxation.}
When the relaxation term is removed, the local dynamics are dominated by the scene-level forcing $\mathbf{W}_i\mathbf{q}(\tau)$ without a Gaussian-specific restoring mechanism. While the degradation within the observed interval is moderate, the extrapolation PSNR drops substantially from 40.98~dB to 38.37~dB, accompanied by the largest LPIPS among all variants. This result shows that the local relaxation term is particularly important for regulating long-term thermal evolution and preventing accumulated prediction errors during temporal extrapolation.

\begin{table}[t]
    \centering
    \caption{Ablation study of the GTSM model configurations on temporal interpolation and extrapolation.}
    \label{tab:gtsm_parameter_ablation}
    \scriptsize
    \setlength{\tabcolsep}{2.2pt}
    \begin{adjustbox}{max width=\columnwidth}
    \begin{tabular}{ccc|ccc|ccc}
        \toprule
        \multirow{2}{*}{$D$}
        & \multirow{2}{*}{$M$}
        & \multirow{2}{*}{Reference}
        & \multicolumn{3}{c|}{Interpolation}
        & \multicolumn{3}{c}{Extrapolation} \\
        \cmidrule(lr){4-6}
        \cmidrule(lr){7-9}
        &
        &
        & PSNR$\uparrow$
        & SSIM$\uparrow$
        & LPIPS$\downarrow$
        & PSNR$\uparrow$
        & SSIM$\uparrow$
        & LPIPS$\downarrow$ \\
        \midrule

        \multicolumn{9}{l}{\textit{Baseline}} \\
        3 & 4 & Midpoint
        & 41.15 & 0.990 & 0.055
        & 40.98 & 0.991 & 0.055 \\

        \midrule
        \multicolumn{9}{l}{\textit{Local thermal state dimension $D$}} \\
        1 & 4 & Midpoint
        & 40.63 & 0.989 & 0.052
        & 39.79 & 0.991 & 0.058 \\
        5 & 4 & Midpoint
        & 41.06 & 0.990 & 0.052
        & 39.99 & 0.991 & 0.056 \\

        \midrule
        \multicolumn{9}{l}{\textit{Scene-level driver dimension $M$}} \\
        3 & 2 & Midpoint
        & 41.07 & 0.988 & 0.057
        & 39.61 & 0.991 & 0.059 \\
        3 & 6 & Midpoint
        & 41.05 & 0.990 & 0.057
        & 39.63 & 0.991 & 0.052 \\

        \midrule
        \multicolumn{9}{l}{\textit{Thermal reference state}} \\
        3 & 4 & First
        & 40.79 & 0.991 & 0.057
        & 38.60 & 0.988 & 0.060 \\
        3 & 4 & Last
        & 40.57 & 0.990 & 0.059
        & 38.36 & 0.989 & 0.062 \\

        \bottomrule
    \end{tabular}
    \end{adjustbox}
\end{table}

Table~\ref{tab:gtsm_parameter_ablation} evaluates the effects of the GTSM state dimensions and thermal reference placement. The dimensions \(D\) and \(M\) control the capacities of the Gaussian-local thermal states and the shared scene-level process, respectively, while the reference position determines the temporal rollout direction: First mainly uses forward integration, Last mainly uses backward integration, and Midpoint uses bidirectional integration from the center of the observed interval. Variations in \(D\) and \(M\) lead to only minor performance differences, whereas the reference placement has a more pronounced influence on extrapolation. Moving the reference from Midpoint to First or Last reduces the extrapolation PSNR from \(40.98\) dB to \(38.60\) and \(38.36\) dB, respectively. Notably, Last performs worse despite its shorter forward rollout distance to the future interval, suggesting that the advantage of Midpoint arises from the more balanced bidirectional constraints provided by observations on both sides of the reference state, rather than from rollout length alone.

\textbf{Efficiency analysis.}
GTSM introduces a compact set of Gaussian-specific thermal parameters,
including $B_i$, $\mathbf{z}_i^r$, $\boldsymbol{\lambda}_i$, and
$\mathbf{W}_i$, corresponding to $1+2D+DM=19$ scalars per Gaussian under
the default setting $D=3$ and $M=4$, together with a lightweight
scene-level driver shared by all Gaussians. In our RHD experiments, GTSM
requires an average GPU memory of approximately 2759~MiB, compared with
8694~MiB for ETGS. However, ETGS evaluates its parameterized thermal response
through a closed-form solution, whereas GTSM numerically propagates the
coupled scene-level and Gaussian-local states using RK2 during optimization.
Consequently, the average training time increases from approximately
655~s for ETGS to 1975~s for GTSM. Thus, GTSM reduces GPU memory consumption
but incurs additional computation for continuous state-space rollout.

\begin{figure}[t]
    \centering
    \includegraphics[width=\columnwidth]{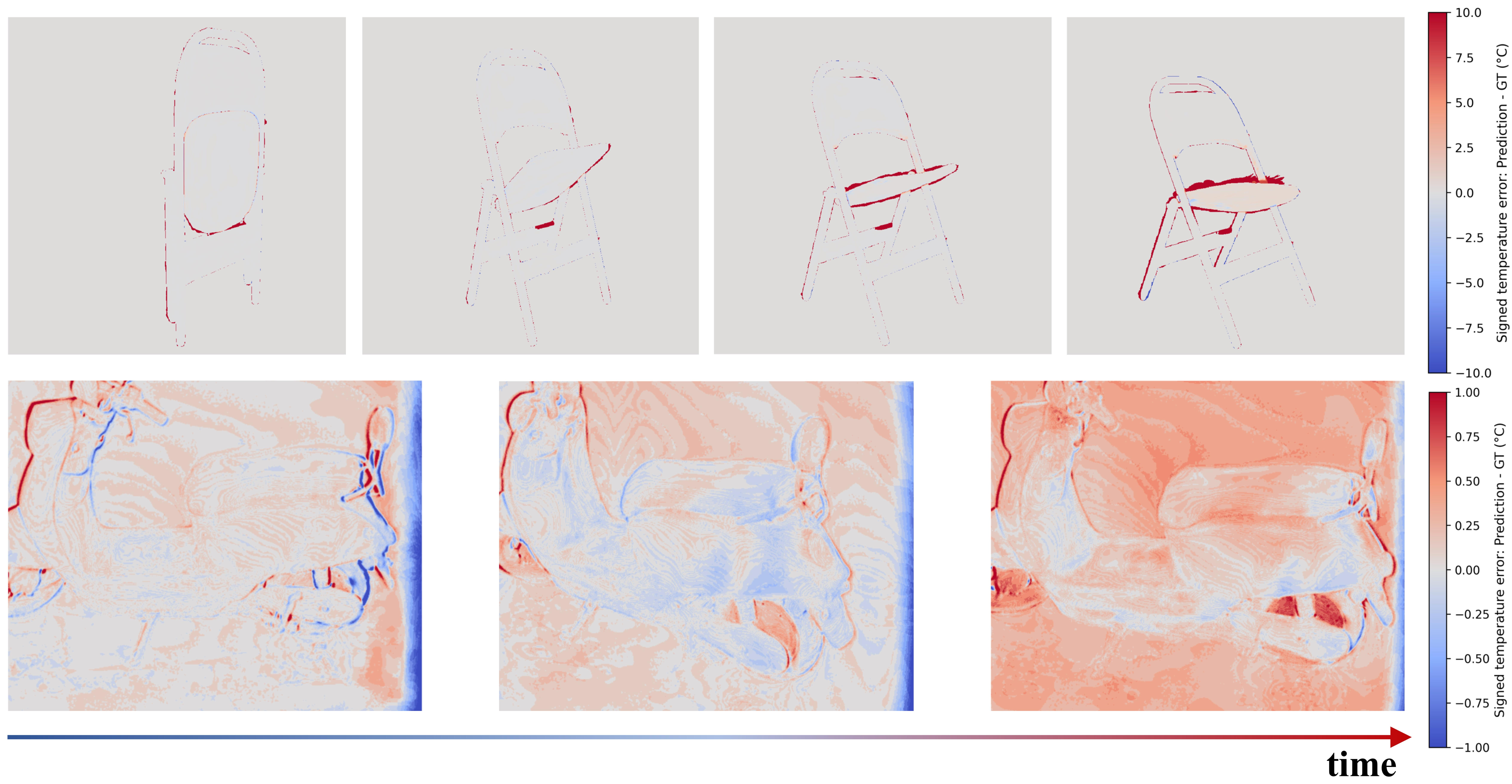}
    \caption{ Visualization of signed temperature prediction errors for the dynamic Foldingchair scene (top) and the static cooling Ebike scene (bottom). Note that different color ranges are used for the two scenes to reveal their respective error structures.
    }
    \label{fig:thermal_error_analysis}
\end{figure}

Fig.~\ref{fig:thermal_error_analysis} further visualizes the spatial distribution of temperature prediction errors for dynamic and static scenes. For the dynamic Foldingchair scene, the prediction remains accurate over most object-interior regions, whereas larger errors increasingly concentrate around moving boundaries at future timestamps. This behavior mainly results from geometric extrapolation errors in the 4DGS backbone: a small spatial misalignment between the predicted and ground-truth geometry can assign the target temperature to background pixels, or vice versa, producing relatively large temperature residuals near object boundaries. In comparison, the static cooling E-bike scene does not accumulate geometric deformation errors. Although the thermal prediction error gradually increases with the extrapolation horizon, it remains small over most regions, indicating that GTSM can maintain accurate future thermal-state prediction when the spatial support is stable.

\section{Conclusion}
\label{sec5}

This paper presents the Gaussian Thermal State-Space Model (GTSM) for reconstructing and predicting four-dimensional spatio-temporal thermal fields from two-dimensional observations. GTSM models continuous Gaussian-level thermal dynamics and can be integrated with both static 3DGS and dynamic 4DGS representations. Experiments on the real RHD dataset and synthetic dynamic thermal data demonstrate accurate thermal-field reconstruction and strong future-time prediction, with particularly notable improvements in temporal extrapolation. The results show the potential of GTSM for thermal-state reconstruction and prediction in industrial monitoring, dynamic target analysis, and predictive infrared perception.


\section*{Conflict of Interest}
The authors declare that there are no known competing financial interests or personal relationships that could have appeared to influence the work reported in this paper.

\bibliographystyle{IEEEtran}
\bibliography{bibliography}

\vfill

\end{document}